\documentclass[11pt,a4paper]{article}
\usepackage[hyperref]{acl2020}
\usepackage{times}
\usepackage{latexsym}
\usepackage{amsmath}
\usepackage{graphicx}

\usepackage{microtype}
\usepackage{enumitem}
\usepackage{booktabs}

\aclfinalcopy %

\title{Considering Likelihood in NLP Classification Explanations with Occlusion and Language Modeling}

\author{{David Harbecke \and Christoph Alt}\\
  German Research Center for Artificial Intelligence (DFKI), Berlin, Germany\\
  \texttt{\{firstname\}.\{lastname\}@dfki.de} \\}

\date{}

\begin{document}
\maketitle

\newcommand{\insertlikelihoodfigure}{
 \begin{figure}[ht]
  \centering
  \includegraphics[trim=80 0 0 0, width=0.45\textwidth]{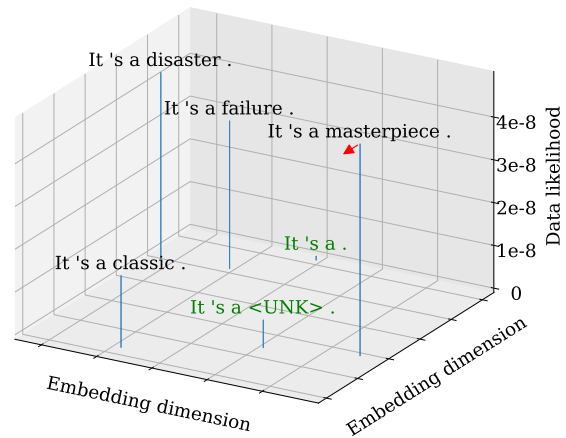}
  \caption{Schematic display of data likelihood in NLP.
  There are discrete inputs, i.e. combination of tokens, with a data likelihood greater than zero.
  All other inputs in the embedding space have likelihood zero because they have no corresponding tokens.
  Occlusion methods (green) create unlikely input.
  Gradient-based explanation methods (red arrow) consider infinitesimal changes to the input and thus data with no likelihood.}
  \label{fig:likelihood_figure}
 \end{figure}
}

\newcommand{\insertexampletable}{
 \begin{table*}[ht]
  \centering
  \begin{tabular}{l|l|l}
    Method&Relevances&Max. value \\ \hline
    \textbf{OLM} (ours)&\colorbox[RGB]{255,133,133}{\strut forced} \colorbox[RGB]{254,254,255}{\strut ,} \colorbox[RGB]{255,238,238}{\strut familiar} \colorbox[RGB]{254,254,255}{\strut and} \colorbox[RGB]{255,224,224}{\strut thoroughly} \colorbox[RGB]{255,0,0}{\strut condescending} \colorbox[RGB]{254,254,255}{\strut .} &0.76\\
    \textbf{OLM-S} (ours)&\colorbox[RGB]{255,0,0}{\strut forced} \colorbox[RGB]{255,254,254}{\strut ,} \colorbox[RGB]{255,136,136}{\strut familiar} \colorbox[RGB]{255,254,254}{\strut and} \colorbox[RGB]{255,98,98}{\strut thoroughly} \colorbox[RGB]{255,27,27}{\strut condescending} \colorbox[RGB]{255,254,254}{\strut .} &0.47\\ \hline
    Delete&\colorbox[RGB]{254,254,255}{\strut forced} \colorbox[RGB]{254,254,255}{\strut ,} \colorbox[RGB]{254,254,255}{\strut familiar} \colorbox[RGB]{254,254,255}{\strut and} \colorbox[RGB]{254,254,255}{\strut thoroughly} \colorbox[RGB]{255,0,0}{\strut condescending} \colorbox[RGB]{254,254,255}{\strut .} &1\\
    UNK&\colorbox[RGB]{254,254,255}{\strut forced} \colorbox[RGB]{254,254,255}{\strut ,} \colorbox[RGB]{254,254,255}{\strut familiar} \colorbox[RGB]{254,254,255}{\strut and} \colorbox[RGB]{255,252,252}{\strut thoroughly} \colorbox[RGB]{255,0,0}{\strut condescending} \colorbox[RGB]{255,254,254}{\strut .} &0.35\\
    Sensitivity Analysis&\colorbox[RGB]{255,128,128}{\strut forced} \colorbox[RGB]{255,196,196}{\strut ,} \colorbox[RGB]{255,172,172}{\strut familiar} \colorbox[RGB]{255,196,196}{\strut and} \colorbox[RGB]{255,159,159}{\strut thoroughly} \colorbox[RGB]{255,0,0}{\strut condescending} \colorbox[RGB]{255,205,205}{\strut .} &0.025\\
    Gradient*Input&\colorbox[RGB]{114,114,255}{\strut forced} \colorbox[RGB]{255,115,115}{\strut ,} \colorbox[RGB]{255,251,251}{\strut familiar} \colorbox[RGB]{255,155,155}{\strut and} \colorbox[RGB]{255,36,36}{\strut thoroughly} \colorbox[RGB]{0,0,255}{\strut condescending} \colorbox[RGB]{255,126,126}{\strut .} &0.00011\\
    Integrated Gradients&\colorbox[RGB]{255,251,251}{\strut forced} \colorbox[RGB]{240,240,255}{\strut ,} \colorbox[RGB]{255,208,208}{\strut familiar} \colorbox[RGB]{241,241,255}{\strut and} \colorbox[RGB]{216,216,255}{\strut thoroughly} \colorbox[RGB]{255,0,0}{\strut condescending} \colorbox[RGB]{255,238,238}{\strut .} &0.68\\
  \end{tabular}
  \caption{Relevance scores of different gradient- and occlusion-based explanation methods for a sentence from the SST-2 dataset, correctly classified as negative sentiment by \emph{RoBERTa}.
  Red indicates an input token, with a contribution to the true label (negative sentiment), blue indicates a detraction from the true label.
  Coloring are normalized for each method for visibility, the maximum value of each method is indicated in the last column.
  The relevances of the first four and last method can be interpreted as prediction difference if that token is missing (see \emph{Sensitivity-1} in \ref{subsec:axioms}).
  The first token ``forced'' only has high relevance for our methods, the most commonly resampled tokens can be found in Table \ref{tab:sampling_examples}.
  Punctuation marks have less relevance than words for our method compared to gradient methods.
  }
  \label{tab:example_explanations}
 \end{table*}
}

\newcommand{\insertsamplingtable}{
 \begin{table}[t]
  \centering
  \small
  \begin{tabular}{l|l|l||l|l|l}
    token&freq.&pred.&token&freq.&pred.\\ \hline
    familiar&9&1&old&2&1\\
    warm&4&7e-4&perfect&2&3.9e-4\\
    ancient&3&0.074&quiet&2&1\\
    cold&3&1&real&2&6.5e-3\\
    beautiful&2&1.4e-4&sweet&2&1.9e-4\\
    bold&2&0.63&wonderful&2&3.1e-4\\
    low&2&1&yes&2&1\\
    nice&2&8.3e-4&young&2&0.99\\
  \end{tabular}
  \caption{Most frequently resampled words for ``forced'' in ``\textit{forced} , familiar and thoroughly condescending .'' from Table \ref{tab:example_explanations} with the prediction of the true label neuron (negative sentiment).
  We sample 100 times per token, the prediction is rounded to two significant digits.
  Many resampled words lead to a positive sentiment classification.
  The high variance of the model prediction for replacements of this token is not captured by another method.}
  \label{tab:sampling_examples}
 \end{table}
}

\newcommand{\insertcombinedcorrelationtable}{
    \begin{table*}[ht]\centering
        \begin{tabular}{@{}lcccccc@{}}
        \toprule
        {} & \multicolumn{2}{c}{MNLI} & \multicolumn{2}{c}{SST-2} & \multicolumn{2}{c}{CoLA} \\
        {} &   OLM & OLM-S &   OLM & OLM-S &  OLM & OLM-S \\
        \midrule
        Delete       &  0.60 &  - &  0.52 &  - & 0.25 &  - \\
        UNK          &  0.58 &  - &  0.47 &  - & 0.21 &  - \\
        Sensitivity Analysis  &  0.27 &  0.35 &  0.30 &  0.37 & 0.20 &  0.29 \\
        Gradient*Input   & -0.03 & - &  0.02 &  - & 0.02 &  - \\
        Integrated Gradients &  0.28 &  - &  0.35 &  - & 0.15 &  - \\
        \bottomrule
        \end{tabular}
    \caption{Correlation between explanation methods on MNLI, SST-2, and CoLA development sets. OLM correlates with every method except for Gradient*Input. The correlation is highest with the other Occlusion methods for MNLI and SST-2 but not close to 1. For all methods, the correlation is lowest on CoLA.}
    \label{tab:explanation_correlation}
    \end{table*}
}

\begin{abstract}
Recently, state-of-the-art NLP models gained an increasing syntactic and semantic understanding of language, and explanation methods are crucial to understand their decisions. Occlusion is a well established method that provides explanations on discrete language data, e.g.\ by removing a language unit from an input and measuring the impact on a model's decision. We argue that current occlusion-based methods often produce invalid or syntactically incorrect language data, neglecting the improved abilities of recent NLP models. Furthermore, gradient-based explanation methods disregard the discrete distribution of data in NLP. Thus, we propose \textbf{OLM}: a novel explanation method that combines \textbf{o}cclusion and \textbf{l}anguage \textbf{m}odels to sample valid and syntactically correct replacements with high likelihood, given the context of the original input.
We lay out a theoretical foundation that alleviates these weaknesses of other explanation methods in NLP and provide results that underline the importance of considering data likelihood in occlusion-based explanation.\footnote{Our experiments are available at \url{https://github.com/DFKI-NLP/OLM}}

\end{abstract}

\insertlikelihoodfigure
\insertexampletable

\section{Introduction}
\label{introduction}
Explanation methods are a useful tool to analyze and understand the decisions made by complex non-linear models, e.g.\ neural networks. For example, they can attribute \emph{relevance} scores to input features (e.g.\ word or sub-word units in NLP).
Gradient-based methods provide explanations by analyzing local infinitesimal changes to determine the shape of a network's function. The implicit assumption is that the local shape of a function is indicative or useful to calculate the relevance of an input feature for a model's prediction. In computer vision, for example, infinitesimal changes to an input image still produce another valid image and the change in prediction is a valid tool to analyze what led to it \citep[e.g.,][]{zintgraf2017visualizing}. The same applies to methods that analyze the function's gradient at multiple points, such as \emph{Integrated Gradients} \cite{sundararajan2017axiomatic}.

In NLP, however, the input consists of natural language, which is discrete, i.e.,\ the data that has positive likelihood is a discrete distribution (see Figure \ref{fig:likelihood_figure}).
This means that local neighborhoods need not be indicative of the model's prediction behaviour and a model's prediction function at points with zero likelihood need not be relevant to the model's decision.
Thus, we argue that black-box models in NLP should be analyzed only at inputs of non-zero likelihood and explanation methods should not rely on gradients. %
Occlusion is a well suited method due to its ability to produce explanations on data with discrete likelihood. For example, by replacing or deleting a language unit in the original input and measuring the impact on the model's prediction.
However, the likelihood of the replacement data is usually low.
Consider, for example, a sentiment classification task and assume a model that assigns syntactically incorrect inputs a negative sentiment. It correctly predicts \emph{``It 's a masterpiece .''} as positive, but assigns negative sentiment to syntactically incorrect inputs produced by occlusion, e.g.\ \emph{``It 's a .''} or \emph{``It 's a $<$UNK$>$ .''}, which have low data likelihood (see Figure \ref{fig:likelihood_figure}).
This may result in a large prediction difference for many tokens in a positive sentiment example and no prediction difference for many tokens in a negative sentiment example (see Table \ref{tab:example_explanations}), independent of whether they carry any sentiment information and thus may be relevant to the model.
This example shows that the relevance attributed by current occlusion-based methods may depend solely on the model's syntactic understanding instead of the input feature's information regarding the task.

We argue that current NLP state-of-the-art models have increasing syntactic \citep{hewitt2019structural} and hierarchical \citep{liu2019linguistic} understanding. 
Therefore, methods that explain these models should consider syntactically correct replacement that is likely given the unit's context, e.g.\ in \mbox{Figure \ref{fig:likelihood_figure}} \textit{``classic''} or \textit{``failure''} as replacements for \textit{``masterpiece''} in \textit{``It 's a masterpiece .''}
Our experiments show that presenting these models with perturbed ungrammatical input changes the explanations.

\subsection{Contributions}
\begin{itemize}[nosep,leftmargin=*]
    \item We present \textbf{OLM}, a novel black-box relevance explanation method which considers syntactic understanding.
    It is suitable for any model that performs an NLP classification task and we analyze which axioms for explanation methods it fulfills.
    \item We introduce the \textbf{class zero-sum axiom} for explanation methods.
    \item We experimentally compare the relevances produced by our method to those of other black-box and gradient-based explanation approaches.
\end{itemize}

\section{Methods}

In this section, we introduce our novel explanation method that combines occlusion with language modeling. Instead of deleting or replacing a linguistic unit in the input with an unlikely replacement, \emph{OLM} substitutes it with one generated by a language model. This produces a contextualized distribution of valid and syntactically likely reference inputs and allows a more faithful analysis of models with increasing syntactic capabilities. This is followed by an axiomatic analysis of \emph{OLM}'s properties. Finally, we introduce \textbf{OLM-S}, an extension that measures sensitivity of a model at a feature's position.

For our approach we employ the difference of probabilities formula from \citet{robnik2008explaining}.
Let $x_i$ be an attribute of input $x$ and $x_{\setminus i}$ the incomplete input without this attribute.
Then the relevance $r$ given the prediction function $f$ and class $c$ is
\begin{equation}
    r_{f, c}(x_i) = f_c(x) - f_c(x_{\setminus i}).
    \label{eqn:occlusion}
\end{equation}
Note that $f_c(x_{\setminus i})$ is not accurately defined and needs to be approximated, as $x_{\setminus i}$ is an incomplete input.
For vision, \citet{zintgraf2017visualizing} approximate $f_c(x_{\setminus i})$ by using the input data distribution $p_{data}$ to sample $\hat{x}_i$ independently of $x$ or use a Gaussian distribution for $\hat{x}_i$ conditioned on surrounding pixels.
We argue sampling should be conditioned on the whole input and depend on the probability of the data distribution.
We argue that in NLP a language model $p_{LM}$ generates input that is as natural as possible for the model and thus approximate
\begin{equation}
    f_c(x_{\setminus i}) \approx \sum_{\hat{x}_i} p_{LM}(\hat{x}_i|x_{\setminus i}) f_c(x_{\setminus i}, \hat{x}_i).
    \label{eqn:lm_approxmation}
\end{equation}
In general, $x_i$ should be units of interest such as phrases, words or subword tokens.
Thus, \emph{OLM}'s relevance for a language unit is the difference in prediction between the original input and inputs with the unit resampled by conditioning on information in its context.
The relevance of every language unit is in the interval $[-1, 1]$, with the sign indicating contradiction or support, and can be interpreted as the value of information added by the unit for the model.

\insertsamplingtable

\subsection{Axiomatic Analysis}
\label{subsec:axioms}
\citet{sundararajan2017axiomatic} introduced axiomatic development and analysis of explanation methods.
We follow their argument that an explanation method should be derived theoretically, not experimentally, as we want to analyze a model, not our understanding of it.
First, we introduce a new axiom.
Then we discuss which existing axioms our method fulfills.\footnote{Proofs for the following analysis can be found in Appendix \ref{appendix}.}

\textbf{Class Zero-Sum Axiom.}
We introduce an axiom that follows from the intuition that for a normalized DNN every input feature contributes as much to a specific class as it detracts from all other classes.
Let $f$ be a prediction function where the output is normalized over all classes $C$.
Every input feature contributes as much to the classification of a specific class as it detracts from other classes.
A relevance method that gives a feature positive relevance for every class is not helpful in understanding the model.
An explanation method satisfies \emph{Class Zero-Sum} if the summed relevance of each input feature $x_i$ over all classes is zero.
\begin{equation}
    \sum_{c \in C} r_{f, c}(x_i) = 0
\end{equation}
This axiom can be seen as an alternative to the \textbf{Completeness} axiom given by \citet{bach2015pixel}.
\emph{Completeness} states that the sum of the relevances of an input is equal to its prediction.
They can not be fulfilled simultaneously.
\citet{gosiewska2019donottrust} show that a linear distribution of relevance as with \emph{Completeness} is not necessarily desirable for non-linear models.
They argue that explanations that force the sum of relevances to be equal to the prediction do not capture the interaction of features faithfully.
\emph{OLM} fulfills \emph{Class Zero-Sum}, as do other occlusion methods and gradient methods.
Other axioms \emph{OLM} fulfills are:

\textbf{Implementation Invariance.} Two neural networks that represent the same function, i.e.~give the same output for each possible input, should receive the same relevances for every input \citep{sundararajan2017axiomatic}.

\textbf{Linearity.} A network, which is a linear combination of other networks, should have explanations which are the same linear combination of the original networks explanations \citep{sundararajan2017axiomatic}.

\textbf{Sensitivity-1.} The relevance of an input variable should be the difference of prediction when the input variable is occluded \citep{ancona2017towards}.

\insertcombinedcorrelationtable

\subsection{OLM-S}
From our approach we can also deduce a method that describes the sensitivity of the classification at the position of an input feature.
To this end, we compute the standard deviation of the language model predictions.
\begin{equation}
    s_{f, c}(x_i) = \sqrt{\sum_{\hat{x}_i} p_{LM}(\hat{x}_i|x_{\setminus i}) \left(f_c(x_{\setminus i}, \hat{x}_i) - \mu \right)^2},
\end{equation}
where $\mu$ is the mean value from equation \ref{eqn:lm_approxmation}.
We call this \emph{OLM-S(ensitivity)}.
Note that this measure is independent of $x_i$ and only describes the sensitivity of the feature's position.
This means that it measures a model's sensitivity at a given language unit's position given the context.
\emph{OLM} and \emph{OLM-S} are thus using mean and standard deviation, respectively, of the prediction when resampling a token.

\section{Experiments}
In our experiments, we aim to answer the following question: Do relevances produced by our method differ from those that either ignore the discrete structure of language data or produce syntactically incorrect input, and if so, how?

We first train a state-of-the-art NLP model \citep[\emph{RoBERTa},][]{liu2019roberta} on three sentence classification tasks (Section~\ref{subsec:tasks}). We then compare the explanations produced by \emph{OLM} and \emph{OLM-S} to five occlusion and gradient-based methods (Section~\ref{subsec:baseline_methods}). To this end, we calculate the relevances of words over a whole input regarding the true label. We calculate the Pearson correlation coefficients of these relevances for every sentence and average this over the whole development set of each task. In our experiments we use \emph{BERT base} \citep{devlin-etal-2019-bert} for \emph{OLM} resampling.

\subsection{Baseline Methods}
\label{subsec:baseline_methods}
We compare \emph{OLM} with occlusion \citep{robnik2008explaining, zintgraf2017visualizing} in two variants.
One method of occlusion is \textbf{deletion} of the word.
The other method is replacing the word with the \textbf{$<$UNK$>$} token for unknown words.
These methods can produce ungrammatical input, as we argue in Section \ref{introduction}.

Furthermore, we compare with the following gradient-based methods.
\textbf{Sensitivity Analysis} \citep{simonyan2013deep} is the absolute value of the gradient.
\textbf{Gradient*Input} \citep{shrikumar2016not} is simple component-wise multiplication of an input with its gradient.
\textbf{Integrated Gradients} \citep{sundararajan2017axiomatic} integrate the gradients from a reference input to the current input.
As these gradient-based methods provide relevance for every word vector value, we sum up all vector values belonging to a word.
Gradient-based methods do not consider likelihood in NLP (see Section \ref{introduction}) and are thus also merely a comparison and not a gold standard.

\subsection{Tasks}
\label{subsec:tasks}
We select a representative set of NLP sentence classification tasks that focus on different aspects of context and linguistic properties:

\textbf{MNLI (matched)}
The Multi-Genre Natural Language Inference Corpus \cite{williams2018broad} contains 400k pairs of premise and hypothesis sentences and the task is to predict whether the premise entails the hypothesis.
We re-use the \emph{RoBERTa large} model fine-tuned on MNLI \cite{liu2019roberta}, with a dev set accuracy of 90.2.

\textbf{SST-2}
The Stanford Sentiment Treebank \cite{socher2013recursive} contains 70k sentences labeled with positive or negative sentiment.
We fine-tune the pre-trained \emph{RoBERTa base} to the classification task and achieve an accuracy of 94.5 on the dev set.

\textbf{CoLA}
The Corpus of Linguistic Acceptability \cite{warstadt2018neural} contains 10k sentences labeled as grammatical or ungrammatical, e.g.~`They can sing.' (acceptable) vs.~ `many evidence was provided.' (unacceptable).
Similar to SST-2, we fine-tune \emph{RoBERTa base} to the task and achieve a Matthew's corr.~ of 61.3 on the dev set.

\subsection{Results}
\label{subsec:results}
Table \ref{tab:explanation_correlation} shows the correlation of our two proposed occlusion methods (\emph{OLM} and \emph{OLM-S}) with other explanation methods on three NLP tasks. For \emph{OLM-S} we only report correlation to \emph{Sensitivity} because both inform about the magnitude of possible change.
They both provide non-negative values and therefore are not necessarily comparable to the other methods.
We find that across all tasks \emph{OLM} correlates the most with the two occlusion-based methods (\emph{Unk} and \emph{Delete}) but the overall correlation is low, with a maximum of 0.6 on MNLI.
Also the level differs greatly between tasks, ranging from 0.21 and 0.25 (\emph{Unk}, \emph{Delete}) on CoLA to 0.58 and 0.6 on MNLI.
As this is an average of correlations, this shows that resampling creates distinctive explanations that can not be approximated by other occlusion methods.
An example input from SST-2 can be found in Table \ref{tab:example_explanations}, which clearly highlights the difference in explanations.
Table \ref{tab:sampling_examples} shows the corresponding tokens resampled by \emph{OLM}, using \emph{BERT base} as the language model.
For gradient-based methods the correlation with \emph{OLM} is even lower, ranging from -0.03 for \emph{Gradient*Input} on MNLI to 0.35 for \emph{Integrated Gradients} on SST-2. For \emph{OLM-S} we observe a correlation between 0.29 (CoLA) and 0.35 (MNLI), which is still low.
\emph{Gradient*Input} shows almost no correlation to \emph{OLM} across tasks.
The overall low correlation of gradient-based methods with \emph{OLM} and \emph{OLM-S} suggests that ignoring the discrete structure of language data might be problematic in NLP.

\section{Related Work}

There exist many other popular black-box explanation methods for DNNs.
\emph{SHAP} \citep{lundberg2017unified} is a framework that uses Shapley Values which are a game-theoretic black-box approach to determining relevance by occluding subsets of all features.
They do not necessarily consider the likelihood of data.
The occlusion \emph{SHAP} employs may be combined with \emph{OLM} but the approximation error of the language model could increase with more features occluded.
\emph{LIME} \citep{ribeiro2016should} explains by learning a local explainable model.
\emph{LIME} tries to be locally faithful to a model, which is, as we argue, not as important as likely data for explanations in NLP.

There are also explanation methods for DNNs which give layer-specific rules to retrieve relevance.
\emph{LRP} \citep{bach2015pixel} propagates relevance from the output to the input such that \emph{Completeness} is satisfied for every layer.
\emph{DeepLIFT} \citep{shrikumar2017learning} compares the activations of an input with activations reference inputs.
In contrast to \emph{OLM}, these layer-specific explanation methods have been shown not to satisfy \emph{Implementation Invariance} \cite{sundararajan2017axiomatic}.

Most state-of-the-art models in NLP are transformers which use attention.
There is a discussion on whether attention weights \citep{bahdanau2014neural, vaswani2017attention} should be considered as explanation method in \citet{jain2019attention} and \citet{wiegreffe2019attention}.
They are not based on an axiomatic attribution of relevances.
It is unclear whether they satisfy any axiom.
An advantage to analyzing attention weights is that attention weights naturally show what the model does.
Thus, even if they do not always provide a faithful explanation, their analysis might be helpful for a specific input.

\section{Conclusion}
We argue that current black-box and gradient-based explanation methods do not yet consider the likelihood of data and present \emph{OLM}, a novel explanation method, which uses a language model to resample occluded words.
It is especially suited for word-level relevance of sentence classification with state-of-the-art NLP models.
We also introduce the \emph{Class Zero-Sum Axiom} for explanation methods, compare it with an existing axiom.
Furthermore, we show other axioms that \emph{OLM} satisfies.
We argue that with this more solid theoretical foundation \emph{OLM} can be regarded as an improvement over existing NLP classification explanation methods.
In our experiments, we compare our methods to other occlusion and gradient explanation methods.
We do not consider these experiments to be exhaustive.
Unfortunately, there is no general evaluation for explanation methods.

We show that our method adds value by showing distinctive results and better founded theory.
A practical difficulty of \emph{OLM} is the approximation with a language model.
First, a language model can create syntactically correct data, that does not make sense for the task.
Second, even state-of-the-art language models do not always produce syntactically correct data.
However, we argue that using a language model is a suitable way for finding reference inputs.

In the future, we want to extend this method to language features other than words.
NLP tasks with longer input are probably not very sensitive to single word occlusion, which could be measured with \emph{OLM-S}.

\section*{Acknowledgments}
We would like to thank Leonhard Hennig, Robert Schwarzenberg and the anonymous reviewers for their feedback on the paper. 
This work was partially supported by the German Federal Ministry of Education and Research as part of the projects BBDC2 (01IS18025E) and XAINES.

\bibliography{acl2020}
\bibliographystyle{acl_natbib}

\appendix
\section{Proof Appendix}
\label{appendix}
Let $f$ be a neural network that predicts a probability distribution over classes $C$, i.e. $\sum_{c\in C}f_c(x) = 1$.
Let $x=(x_1,...,x_n)$ be a input split into $n$ input features. %

1. \textbf{Class Zero-Sum} and \textbf{Completeness} rule each other out. Assume $r_{f,c}$ fulfills both, then we have
\begin{equation}
    \sum_{i=1}^{n}\sum_{c\in C}r_{f,c}(x_i) = 0
\end{equation}
from \textbf{Class Zero-Sum} and 
\begin{equation}
    \sum_{c\in C}\sum_{i=1}^{n}r_{f,c}(x_i) = 1
\end{equation}
from \emph{Completeness}. Contradiction.

2. \textbf{OLM} satisfies \textbf{Class Zero-Sum}.
Let $r_{f,c}$ now be the \emph{OLM} relevance method from equations (1) and (2) in the paper.
\begin{equation}
    \begin{split}
        & \sum_{c\in C}r_{f,c}(x_i) \\
        =& \sum_{c\in C} \left(f_c(x) - \sum_{\hat{x}_i} p_{LM}(\hat{x}_i|x_{\setminus i}) f_c(x_{\setminus i}, \hat{x}_i)\right) \\
        =& \sum_{c\in C} f_c(x) - \sum_{\hat{x}_i} p_{LM}(\hat{x}_i|x_{\setminus i}) \sum_{c\in C} f_c(x_{\setminus i}, \hat{x}_i)\\
        =& 1 - \sum_{\hat{x}_i} p_{LM}(\hat{x}_i|x_{\setminus i}) = 0.
    \end{split}
\end{equation}

3. \textbf{OLM} satisfies \textbf{Implementation Invariance}.
\emph{OLM} is a black box method and only evaluates the function of the neural network.
Thus, it has to satisfy \emph{Implementation Invariance}.

4. \textbf{OLM} satisfies \textbf{Sensitivity-1}.
\emph{OLM} is defined as an Occlusion method, so it necessarily gives the difference of prediction when an input variable is occluded.

5. \textbf{OLM} satisfies \textbf{Linearity}.
Let $f = \sum_{j=1}^{n} \alpha_j g^j$ be a linear combination of models. Then we have
\begin{equation}
    \begin{split}
        r_{f,c}(x_i)=& f_c(x) - \sum_{\hat{x}_i} p_{LM}(\hat{x}_i|x_{\setminus i}) f_c(x_{\setminus i}, \hat{x}_i) \\
        =& \sum_{j=1}^n \alpha_j g_c^j(x) -\\
        &\quad\sum_{\hat{x}_i} p_{LM}(\hat{x}_i|x_{\setminus i}) \sum_{j=1}^n \alpha_j g_c^j(x_{\setminus i}, \hat{x}_i)\\
        =& \sum_{j=1}^n \alpha_j r_{g^j,c}(x_i).
    \end{split}
\end{equation}

\end{document}